\documentclass[letterpaper,journal]{IEEEtran}
\usepackage{amsmath,amsfonts}
\usepackage{algorithmic}
\usepackage{algorithm}
\usepackage{array}
\usepackage[caption=false,font=normalsize,labelfont=sf,textfont=sf]{subfig}
\usepackage{textcomp}
\usepackage{xcolor}
\usepackage{stfloats}
\usepackage{url}
\usepackage{verbatim}
\usepackage{graphicx}
\usepackage{cite}
\usepackage{graphicx}
\usepackage{float} 
\usepackage{hyperref}
\usepackage{cleveref}
\usepackage[utf8]{inputenc}
\usepackage{orcidlink}
\usepackage{booktabs}
\usepackage[table]{xcolor}
\usepackage{multirow}

\begin{document}

\title{Con-DSO: Learning Short-Horizon Consistency Priors for RGB-D Direct Sparse Odometry}

\author{Haolan Zhang~\orcidlink{0009-0007-1742-3754}, \IEEEmembership{Student member,~IEEE}, Thanh Nguyen Canh~\orcidlink{0000-0001-6332-1002},~\IEEEmembership{Graduate Student member, IEEE}, Chenghao Li~\orcidlink{0009-0003-0404-3825}, \IEEEmembership{Member,~IEEE}, Ziyan Gao~\orcidlink{0000-0001-9948-7960}, \IEEEmembership{Member,~IEEE}, Xiongwen Jiang~\orcidlink{0000-0002-7054-018X}, Nak Young Chong~\orcidlink{0000-0001-5736-0769}, \IEEEmembership{Senior member, IEEE}
\thanks{}
\thanks{Haolan Zhang, Thanh Nguyen Canh, Chenghao Li, Ziyan Gao and Nak Young Chong are with the School of Information Science, Japan Advanced Institute of Science and Technology, Nomi 923-1211, Japan (email: \{haolan.z; thanhnc; chenghao.li; ziyan-g; nakyoung\}@jaist.ac.jp).}
\thanks{Xiongwen Jiang is with College of Information Engineering, Shenyang University of Chemical Technology, Shenyang 110142, China (email: jiang@syuct.edu.cn).}}


\markboth{}%
{Shell \MakeLowercase{\textit{et al.}}: A Sample Article Using IEEEtran.cls for IEEE Journals}


\maketitle

\begin{abstract}
RGB-D direct visual odometry (VO) benefits from metric depth measurements but often degrades in the presence of dynamic objects, occlusions, illumination changes, and unreliable depth, which violate the photometric and geometric consistency assumptions of direct alignment. We propose Con-DSO, a consistency-aware RGB-D direct sparse odometry framework that addresses these challenges through a unified learned uncertainty model. A dual-branch consistency network is trained on adjacent RGB-D frame pairs using flow-guided photometric errors and projective depth-consistency errors to predict pixel-level photometric and geometric uncertainty. The predicted uncertainty is first converted into pairwise quality to guide support-pixel selection and is then fused across adjacent frame pairs to form a host-side quality prior for keyframe-based tracking. To account for the different roles of photometric and depth information in direct RGB-D optimization, the quality prior is incorporated through a decoupled photometric-geometric weighting scheme, with the geometric weight applied only to the translational component of pose estimation. Experiments on five public RGB-D benchmarks demonstrate consistent improvements over direct RGB-D odometry baselines, achieving more than 20\% reduction in absolute trajectory error on ICL-NUIM and approximately 50\% to 80\% reductions on RGB-D Scenes V2, TUM/BONN, and OpenLORIS. These results demonstrate that learned consistency-aware uncertainty can substantially improve the robustness of RGB-D direct visual odometry in challenging environments.

\end{abstract}

\begin{IEEEkeywords}
RGB-D visual odometry, direct sparse odometry, consistency-aware tracking, uncertainty estimation, uncertainty-aware pixel selection, photometric-geometric weighting.
\end{IEEEkeywords}

\section{Introduction}
\IEEEPARstart{V}{isual} odometry (VO) is a fundamental component of many simultaneous localization and mapping (SLAM) systems and plays an important role in robotics, augmented reality, and many other perception applications~\cite{robot,AR,survey1}. Owing to the direct availability of metric depth measurements, RGB-D sensing enables accurate 3D reconstruction and alleviates the scale ambiguity inherent in monocular systems. Over the past decade, visual odometry and SLAM systems have evolved from classical geometric formulations, such as KinectFusion~\cite{KinectFusion}, robust RGB-D odometry~\cite{Robust VO}, ElasticFusion~\cite{ElasticFusion}, and the ORB-SLAM family~\cite{ORB-SLAM,ORB-SLAM2,ORB-SLAM3}, to modern learning-based and neural scene representation approaches such as DROID-SLAM~\cite{DROID} and NICE-SLAM~\cite{NICE}.
However, most existing methods rely, either explicitly or implicitly, on local photometric and depth-geometric consistency across views. In real-world scenes, this assumption is often violated by dynamic objects, occlusion, illumination variation, low-texture regions, and locally unreliable depth. As a result, visual alignment becomes unstable and residual errors increase, significantly degrading the accuracy and robustness of pose estimation.

To improve robustness in the presence of such violations, previous studies have introduced explicit processing mechanisms, including semantic filtering of dynamic objects, occlusion-aware processing, illumination compensation, geometric consistency verification, point-cloud-based outlier rejection, and depth-aware tracking~\cite{DS-SLAM,SG-SLAM,Dynamic-Dso, Blitz-SLAM, occlusion,occlusion-aware,Illumination-Adaptive,CVO,ACVO,Kumar,RGBD-DSO}.

While these approaches can effectively suppress unreliable observations under challenging conditions, those based on external modules typically target a particular source of inconsistency, whereas those relying on explicit consistency criteria usually evaluate a single type of consistency, most commonly geometric consistency. In contrast, within an RGB-D direct odometry formulation, photometric and geometric violations from diverse sources all affect the same photometric residual, which motivates a unified representation of consistency violations.

Compared with explicit processing strategies, another line of research improves robustness by estimating the quality of observations and incorporating the resulting confidence, uncertainty, or informativeness measures into the VO/SLAM pipeline. Existing methods assign static point-level weights or exploit temporal point correlations for dynamic scenes~\cite{point weighting,point correlations}, assess feature quality for feature selection~\cite{Good features,feature selection}, estimate object-level uncertainty~\cite{switching}, employ temporal motion-residual priors for adaptive tracking~\cite{RigidFusion}, combine multiple hand-crafted metrics to guide tracking and optimization~\cite{QualiSLAM,SR-SLAM}, or propagate learned uncertainty into per-feature covariances~\cite{MAC-VO}. Collectively, these studies demonstrate that observation quality assessment can substantially enhance robustness. Despite their differences, these methods share a common characteristic: observation reliability is represented by a single quantity for each observation, such as a confidence score, a selection decision, or a covariance estimate. This reliability measure is then applied uniformly to the entire observation. Such a formulation is well suited to feature-based representations, where observations are naturally defined as sparse features, landmarks, or positions. However, it may be less appropriate for RGB-D direct odometry, where photometric and geometric information contribute through different components of the optimization and may exhibit distinct reliability characteristics.

Motivated by these observations, we propose Con-DSO, a consistency-aware RGB-D direct visual odometry framework built around a learned consistency network. Given a pair of temporally adjacent RGB-D frames, the consistency network predicts two dense uncertainty maps that characterize photometric inconsistency and depth-geometric inconsistency, respectively. The network is trained with a heteroscedastic pixel-wise uncertainty objective on photometric and depth-geometric prediction errors, thereby assigning high uncertainty to inconsistent observations regardless of whether they arise from dynamic objects, occlusions, illumination changes, or unreliable depth.

The predicted consistency is integrated into the VO pipeline through two stages. In the first stage, the pairwise quality guides support-pixel selection, preferentially retaining pixels that are both informative and reliable. In the second stage, the pairwise qualities of the two frame pairs around a keyframe, spanning frames $T{-}1$, $T$, and $T{+}1$, are fused into a host-side quality prior, which reweights the pose linearization through a decoupled photometric-geometric scheme. Across both stages, the photometric and geometric predictions represent diverse challenging factors in a unified manner, and Section~\ref{sec:threshold analysis} compares the proposed approach against direct RGB-D methods based on hand-crafted consistency criteria across five benchmark datasets. In the second stage, the decoupled weighting follows the direct RGB-D formulation, in which photometric error enters the residual directly, whereas depth error enters through the warp and is coupled only to the translational term of the linearization. The photometric weight therefore acts on both the rotational and translational terms, and the geometric weight on the translational term alone. The effectiveness of the proposed decoupled weighting scheme is evaluated in Section~\ref{sec:ablation} against a single shared weight, which cannot suppress unreliable depth without also weakening informative image-gradient constraints.

Our main contributions are as follows:
\begin{enumerate}
    \item A dual-branch consistency network that represents consistency violations from diverse sources in a unified manner as pixel-level photometric and depth-geometric uncertainty, predicted from adjacent RGB-D frame pairs and trained with a heteroscedastic objective on flow-guided photometric and projective depth-consistency errors.

    \item A two-stage integration of learned consistency into the VO pipeline: pairwise quality for support-pixel selection, and its bidirectional fusion into a host-side quality prior for keyframe-based tracking.

    \item A decoupled photometric-geometric weighting scheme for direct pose estimation, in which the photometric weight is applied to the full pose linearization, while the geometric weight influences only its translational component.

    \item Experiments on five public RGB-D datasets showing consistent improvements over direct RGB-D baselines, with the network trained only on synthetic TartanAir data and applied without fine-tuning.
\end{enumerate}

The remainder of the paper is organized as follows. Section.~\uppercase\expandafter{\romannumeral2} reviews related work on visual odometry and SLAM under challenging environments. Section.~\uppercase\expandafter{\romannumeral3} presents the Con-DSO framework in detail. Section.~\uppercase\expandafter{\romannumeral4} provides experimental evaluation and analysis. Finally, Section.~\uppercase\expandafter{\romannumeral5} concludes the paper and discusses future work.

\section{Related Work}

Con-DSO improves the robustness of RGB-D direct visual odometry by learning consistency-aware uncertainty and incorporating it into both support-pixel selection and pose estimation. This section reviews two related research directions: explicit robustness-enhancement techniques and observation-quality assessment methods for VO/SLAM.

\subsection{Explicit Robustness Processing in VO/SLAM}

To handle such degradation, many methods introduce explicit robustness processing. Dynamic-scene SLAM methods~\cite{DS-SLAM,SG-SLAM,Dynamic-Dso, Blitz-SLAM} suppress moving regions using semantic segmentation or multi-view geometry constraints~\cite{Multiple View Geometry}. Beyond dynamic-object handling, prior RGB-D odometry and SLAM studies have improved robustness through degradation-specific or geometry-based strategies, including occlusion-aware depth gating or modeling~\cite{occlusion-aware,RGBD-DSO}, illumination-adaptive threshold for feature extraction~\cite{Illumination-Adaptive}, geometric consistency constraints ~\cite{CVO,ACVO,Kumar}.

These approaches generally either target a specific source of consistency violation or evaluate a particular form of consistency using explicit criteria. Semantic filtering methods~\cite{DS-SLAM,SG-SLAM,Dynamic-Dso,Blitz-SLAM} depend on pretrained semantic models and predefined dynamic-object categories. Occlusion and illumination handling techniques identify valid observations using hand-crafted rules tailored to the underlying failure mode, such as depth-difference gating~\cite{occlusion-aware,RGBD-DSO} or adaptive feature-extraction thresholds~\cite{Illumination-Adaptive}. Point-cloud registration methods determine observation validity through parameters including kernel length-scales and sparsification thresholds~\cite{CVO}, while the subsequent ACVO extension~\cite{ACVO} adapts the kernel length-scale online using a manually designed update strategy. Kumar {\it et al.}~\cite{Kumar} formulate the 3D distance consistency of feature correspondences as an image-domain constraint.

\subsection{Observation Quality Assessment in VO/SLAM}

Another line of work evaluates the quality of observations and uses the assessment results to guide subsequent VO/SLAM processing, rather than directly handling challenging factors through external modules or explicit removal strategies. In feature-based methods, feature quality is often defined according to its contribution to pose estimation~\cite{Good features}. Feature-selection methods~\cite{feature selection} evaluate and rank candidate features according to their expected contribution to motion-estimation accuracy. At the object level, Switching~\cite{switching} estimates object quality from feature uncertainty, detection confidence, and object priors, and uses this estimate to determine whether an object enters tightly or loosely coupled pose estimation.

Scene and motion level assessment has also been explored. Joint VO-SF and StaticFusion~\cite{Joint-VO-SF,StaticFusion} estimate static probabilities using K-means clustering to separate reliable static observations from dynamic components. RigidFusion~\cite{RigidFusion} extends this idea by exploiting temporal consistency, using the previous motion segmentation as a prior and evaluating static and dynamic residuals for the current frame. Multi-metric assessment methods further adapt different components of the VO/SLAM pipeline. QualiSLAM~\cite{QualiSLAM} evaluates local blocks with multiple quality indicators, such as illumination variation, motion blur, and feature sparsity, to adapt feature weights in optimization. SR-SLAM~\cite{SR-SLAM}, \textbf{our prior work}, combines detection quality, depth consistency, feature quality, and motion consistency with historical reference frames to guide dynamic handling, keyframe selection, and  optimization weighting. In stereo VO, MAC-VO~\cite{MAC-VO} proposes a learning-based uncertainty assessment by learning the pixel-level uncertainty of optical-flow and stereo-disparity predictions produced by a trained estimation network. The predicted uncertainty serves as a reliability measure for flow-based correspondences used in tracking. High-uncertainty regions are treated as challenging observations, unreliable matches are removed using a relative threshold, and the uncertainty of retained correspondences is propagated into per-keypoint covariance estimates that modulate residual weighting during optimization.

These studies demonstrate that observation-quality assessment can substantially improve robustness. However, in all cases, the assessment outcome is represented by a single quantity per observation, such as a confidence weight, a selection decision, or a covariance estimate, and is applied uniformly to the observation as a whole.

In contrast, Con-DSO learns a pairwise uncertainty from adjacent RGB-D frame pairs that represents dynamic objects, occlusions, illumination changes, and unreliable depth in a unified manner, without dedicated modules or hand-crafted rules for individual sources, and uses the same network configuration on all benchmarks. The resulting pairwise quality guides pixel selection and is fused into a host-side quality prior for the pose estimation. Unlike feature- or position-based observations, the residual in direct RGB-D odometry is a photometric scalar. Photometric and depth-geometric errors contribute to this residual through different terms, and a single reliability quantity per observation cannot characterize both. The quality prior is therefore applied to pose estimation through decoupled photometric and geometric weights.

\section{METHODS}

This section presents the proposed Con-DSO framework. Sec.~\ref{sec:overview} gives an overview of the system pipeline, and Sec.~\ref{sec:network} describes the consistency network. Sec.~\ref{sec:tracking} presents its integration into keyframe (KF) based tracking, and Sec.~\ref{sec:optimization} summarizes the sliding-window optimization inherited from RGBD-DSO~\cite{RGBD-DSO}.

\subsection{Overview of the Proposed System}
\label{sec:overview}

We build Con-DSO upon RGBD-DSO as our foundation. The overall system is illustrated in Fig.~\ref{fig:Pipeline}. Given two temporally adjacent RGB-D frames, our consistency network, trained with flow-guided photometric and depth-projection consistency errors, predicts photometric and geometric (depth) consistency uncertainty. These predictions are first converted into pairwise quality via median normalization. When a keyframe is created, the pairwise quality of the pair ending at it drives pixel selection. Once the following frame arrives, the pair starting at the keyframe is converted in the same way, and the two pairwise qualities are fused bidirectionally at each pixel location into the absolute host-side quality prior attached to the host KF for direct tracking.

In RGBD-DSO, direct tracking is performed by aligning each incoming frame against a reference KF, which serves as the host frame providing reference pixels and depth. Our uncertainty and quality prior are therefore stored on each KF and into KF-to-current-frame tracking in two stages: uncertainty-aware pixel selection and decoupled photometric-geometric (depth) weighting for coarse tracking. In the first stage, the predicted pairwise photometric quality is integrated with the image gradient criterion to guide pixel selection, reducing the likelihood that pixels with high photometric uncertainty are selected as support pixels for direct matching. In the second stage, the photometric and geometric components of the fused quality prior modulate different parts of the coarse tracking Jacobian in a decoupled manner when aligning the current frame to the KF. After these steps, the tracked keyframes are passed to the standard sliding-window optimizer for joint pose and structure refinement.

\begin{figure*}[!htbp]
  \begin{center}
  \includegraphics[width= 0.8\linewidth]{Pipeline.jpg}
  \caption{Overview of Con-DSO. The consistency network predicts photometric and geometric uncertainty from adjacent RGB-D pairs. The resulting pairwise quality guides pixel selection, and its bidirectional fusion into a host-side quality prior drives decoupled weighting during direct tracking, followed by sliding-window optimization.}
  \label{fig:Pipeline}
  \end{center}
\end{figure*}

\subsection{Consistency Network Training}
\label{sec:network}

The consistency network takes two temporally adjacent RGB-D frames $(I_j, D_j)$ and $(I_i, D_i)$ as input and predicts two dense per-pixel uncertainty maps, characterizing the photometric and geometric uncertainty of each pixel, respectively.

We trained our network on the TartanAir dataset~\cite{TartanAir}. Although TartanAir provides stereo imagery, we use only the left-camera RGB images and the associated left-camera depth maps, treating each frame as a single-view RGB-D observation. Ground-truth (GT) optical flow and GT relative poses are used solely as supervision signals to compute reference consistency errors during training. At the inference stage, the network takes raw RGB-D pair inputs only.

\subsubsection{Network \& Training}

Network architecture separates image-driven photometric uncertainty from depth-aware geometric uncertainty, so that the predicted priors can later be integrated into the decoupled pose-estimation scheme. The upper part of Fig.~\ref{fig:Network} illustrates the network structure. The network follows a dual-branch encoder-decoder design. A shared image encoder extracts multi-scale features from both frames at $1/2$, $1/4$, and $1/8$ resolution, and a lightweight depth encoder with the same topology but reduced channel width processes the corresponding depth maps. At the coarsest $1/8$ level, a RAFT-style local correlation volume~\cite{RAFT} is constructed between the two image feature maps using a search radius of $r{=}4$, yielding a $(2r{+}1)^2$-channel matching descriptor. This descriptor provides inter-frame matching evidence for identifying regions where photometric consistency is violated. The geometric branch concatenates this descriptor with image and depth features from both frames, whereas the photometric branch combines it only with image features, excluding depth information. This asymmetric input design prevents depth noise from being conflated with photometric uncertainty. Each branch recovers a full-resolution single-channel uncertainty map through three progressive $2{\times}$ upsampling stages with no final activation, producing the predicted photometric and geometric uncertainty maps, $\hat{\ell}^{\mathrm{photo}}(\mathbf{p})$ and $\hat{\ell}^{\mathrm{geo}}(\mathbf{p})$, respectively, which are parameterized as log-scale values.

\begin{figure}[!htbp]
  \begin{center}
  \includegraphics[width= 0.8\linewidth]{Network.jpg}
  \caption{Overview of the proposed consistency network and training pipeline. Given two temporally adjacent RGB-D frames, the network predicts dense photometric and geometric uncertainty maps through a dual-branch encoder-decoder architecture. Reference photometric and geometric consistency errors are computed using GT optical flow and GT relative pose from TartanAir, detached from the computation graph, and used to train the network with a heteroscedastic NLL loss.}
  \label{fig:Network}
  \end{center}
\end{figure}

The lower part of Fig.~\ref{fig:Network} shows how the ground-truth reference consistency errors are computed. The photometric error $e_{\mathrm{photo}}(\mathbf{p})$ is the $\ell_1$ difference between the current frame and the reference frame warped by GT optical flow, 
\begin{equation}
\label{eq:photo_error}
  e_{\mathrm{photo}}(\mathbf{p})
  = \bigl\|\mathcal{W}(I_j,\mathbf{f}_{j\to i})(\mathbf{p})
    - I_i(\mathbf{p})\bigr\|_1,
\end{equation}
where $\mathcal{W}(\cdot)$ denotes flow-guided image warping, $\mathbf{f}_{j\to i}$ is the GT optical flow from frame $j$ to frame $i$, and $\|\cdot\|_1$ denotes the $\ell_1$ norm averaged over the three color channels of intensities normalized to $[0,1]$.

The geometric error $e_{\mathrm{geo}}(\mathbf{p})$ is the relative depth inconsistency obtained by projecting $D_j$ into the current frame via the GT pose. Let $\mathbf{p}'=\pi(T_{\mathrm{rel}},D_j,K;\mathbf{p})$ denote the projected pixel location in frame $i$, where $\pi(\cdot)$ denotes the pixel projection induced by the projective depth transform using the GT relative pose $T_{\mathrm{rel}}$ and the camera intrinsic matrix $K$. Also let $d_{\mathrm{proj}}(\mathbf{p})$ denote the depth of the transformed 3D pixel. It is then defined as
\begin{equation}
\label{eq:geo_error}
  e_{\mathrm{geo}}(\mathbf{p})
  =
  \frac{
  \bigl|d_{\mathrm{proj}}(\mathbf{p}) - D_i(\mathbf{p}')\bigr|
  }{
  D_i(\mathbf{p}')+\epsilon
  },
\end{equation}
where $D_i(\mathbf{p}')$ is the measured depth sampled at the projected location, and $\epsilon$ is a small constant for numerical stability.

Both GT reference consistency errors are detached from the computation graph before training, so that gradients do not flow through the error computation. The predicted uncertainty maps together with the detached errors are then passed to a heteroscedastic NLL objective under a Laplacian noise assumption. Each loss is evaluated only at pixels for which its corresponding reference error is defined, namely pixels with a valid warp for the photometric term and pixels with a valid depth measurement for the geometric term. The per-pixel losses are defined as

\begin{equation}
\label{eq:loss_photo}
\mathcal{L}_{\mathrm{photo}}(\mathbf{p})
= e_{\mathrm{photo}}(\mathbf{p})\,e^{-\hat{\ell}^{\mathrm{photo}}(\mathbf{p})}
+ \hat{\ell}^{\mathrm{photo}}(\mathbf{p}),
\end{equation}

\begin{equation}
\label{eq:loss_geo}
\mathcal{L}_{\mathrm{geo}}(\mathbf{p})
= e_{\mathrm{geo}}(\mathbf{p})\,e^{-\hat{\ell}^{\mathrm{geo}}(\mathbf{p})}
+ \hat{\ell}^{\mathrm{geo}}(\mathbf{p}),
\end{equation}

where $\hat{\ell}$ is unbounded because the uncertainty heads have no final activation function, and $e^{\hat{\ell}}$ represents the predicted scale. Consequently, the first term in each loss corresponds to the reference error normalized by the scale learned for its respective branch, yielding a dimensionless quantity. As a result, the intensity units in Eq.~\ref{eq:photo_error} and the relative-depth units in Eq.~\ref{eq:geo_error} do not directly affect the combined objective, preventing either branch from dominating the optimization.

At the single image scale, the total loss is 

\begin{equation}
  \mathcal{L} =
    \sum_{\mathbf{p}}
    \Bigl[
      \mathcal{L}_{\mathrm{photo}}(\mathbf{p})
      + \mathcal{L}_{\mathrm{geo}}(\mathbf{p})
    \Bigr].
\end{equation}

The network predicts uncertainty maps only at the full image resolution. Multi-scale supervision is used exclusively during training: the predicted uncertainty maps and corresponding reference errors are average-pooled to scales of $1\times$, $1/2$ and $1/4$ resolution, with weights of 1.0, 0.5, and 0.25, respectively. This auxiliary supervision encourages the predicted uncertainty to capture not only local pixel-level errors but also broader region-level inconsistencies.

The two terms in each per-pixel loss act in opposition: a large scale reduces the first term but increases the second, forcing the network to predict uncertainty that is neither over-confident nor over-uncertain, but matched to the expected residual magnitude.

\subsubsection{Training setup}
The model is trained on TartanAir using 111 trajectories across 8 diverse synthetic environments, with 30 held-out trajectories
for validation. We use AdamW ($\mathrm{lr}{=}8{\times}10^{-4}$, weight decay $10^{-4}$) with a OneCycleLR schedule, 
batch size of 32, and mixed precision training. The final model is trained on a single NVIDIA RTX 3090 (24GB) for approximately 135.5k iterations over 75 hours.

Qualitative evaluation of the learned uncertainty maps is provided in Sec.~\ref{sec:Learned Uncertainty}, demonstrating that the network responds appropriately to diverse challenging conditions.

\subsection{Con-DSO Construction}
\label{sec:tracking}

The learned consistency prior is incorporated into Con-DSO through two stages. First, the network prediction for the frame pair ending at a newly created keyframe is converted into a pairwise quality score, which biases support-pixel selection during keyframe creation. Second, once the subsequent frame arrives, the pairwise qualities of the two frame pairs adjacent to the keyframe are fused to form a host-side quality prior. This prior is then used to reweight the selected pixels during coarse-to-fine direct pose estimation through a decoupled photometric-geometric weighting scheme. 

\subsubsection{Pairwise Quality}
\label{sec:prior generation}

The consistency network described in Sec.~\ref{sec:network} is trained on adjacent frame pairs and predicts uncertainty maps, $\hat{\ell}^k_{t-1,t}$, that parameterize the error scale in logarithmic form. The corresponding scales are given by $\sigma_p=\exp(\hat{\ell}^{\mathrm{photo}})$ for photometric error and $s_{\mathrm{geo}}=\exp(\hat{\ell}^{\mathrm{geo}})$ for relative depth error. They are converted into a pairwise quality map for each pixel location $\mathbf{p}$:
\begin{equation}
\label{eq:pairwise_quality}
Q^k_{t-1, t}(\mathbf{p}) =
\operatorname{clip}
\left(
\frac{m^k_{t-1,t}}
{\exp(\hat{\ell}^k_{t-1,t}(\mathbf{p}))},
\epsilon_q, 1
\right),
\end{equation}
where
$
m^k_{t-1,t}
=
\operatorname{median}_{\mathbf{p}}
\bigl(\exp(\hat{\ell}^k_{t-1,t})\bigr),
$
is the spatial median of the predicted uncertainty map, $k\in\{\mathrm{photo},\mathrm{geo}\}$ denotes the photometric or geometric branch, and $\epsilon_q=10^{-4}$ is a small lower bound used to avoid invalid logarithms in the subsequent log-quality fusion. $Q^k$ describes the consistency of one adjacent transition at pixel location $\mathbf{p}$. This median normalization converts uncertainty into a quality score while remaining robust to extreme predictions.

\subsubsection{Uncertainty-Aware Pixel Selection}
\label{sec:selection}

When a new keyframe $T$ is created, the photometric uncertainty predicted for the frame pair $(T{-}1,T)$ is converted into a pairwise quality score via Eq.~\ref{eq:pairwise_quality}. This score biases support-pixel selection by modulating the original gradient-based pixel score. In the original pixel selector, each candidate pixel is scored according to its directional gradient magnitude along a randomly sampled unit direction:

\begin{equation}
\label{eq:grad_score}
s(\mathbf{p}) = \bigl|\nabla I_T(\mathbf{p}) \cdot \mathbf{d}\bigr|,
\end{equation}
where $\nabla I_T(\mathbf{p}) = (I_x, I_y)^\top$ denotes the image gradient at pixel location $\mathbf{p}$ in keyframe $T$, and $\mathbf{d}$ is a randomly sampled unit direction. Sampling the direction randomly prevents the selected pixels from concentrating along a single edge orientation. Both the scoring function and the sampling strategy are adopted unchanged from the pixel selector of~\cite{RGBD-DSO}.

The pairwise quality $Q^{\mathrm{photo}}_{T-1,T}$ then modulates this score: 

\begin{equation}
\label{eq:pixel_selection}
\tilde{s}(\mathbf{p}) = s(\mathbf{p})\, Q^{\mathrm{photo}}_{T-1,T}(\mathbf{p}).
\end{equation}

Support pixels are then selected by ranking the modulated scores $\tilde{s}(\mathbf{p})$ among candidates satisfying the adaptive threshold criterion $\theta$:

\begin{equation}
\label{eq:ranking}
\mathcal{P}^{*} 
= 
\operatorname*{Top\text{-}K}_{\mathbf{p} \in \mathcal{B}\;
s(\mathbf{p}) > \theta}\,
\tilde{s}(\mathbf{p}),
\end{equation}
where $\mathcal{B}$ denotes the multi-scale candidate grid, $\theta$ is computed as a per-block gradient-magnitude quantile plus a constant offset, and $K$ is the number of retained pixels, following the original RGBD-DSO pixel-selection strategy. The Top-$K$ operation is applied independently within each block rather than globally. Random directional sampling primarily affects which pixel is selected within a block, rather than excluding well-conditioned pixels from consideration. Consequently, the support-pixel budget remains unchanged from the baseline, while the ranking is biased toward pixels that are both informative and temporally reliable.

\subsubsection{Bidirectional Host-side Quality Prior}
\label{sec:fusion}

The pairwise qualities provide short-horizon consistency estimates for adjacent frame pairs. However, RGBD-DSO performs direct tracking between a host keyframe and the current frame, rather than between temporally adjacent frames. To bridge this mismatch, we convert the pairwise qualities into an absolute host-side prior associated with each keyframe, approximating consistency over the entire tracking window. The pairwise quality $Q^{k}_{T,T+1}$ becomes available only after frame $T{+}1$ arrives, by which time the support pixels of keyframe $T$ have already been selected using $Q^{\mathrm{photo}}_{T-1,T}$. Once $Q^{k}_{T,T+1}$ is obtained, the two pairwise qualities adjacent to the keyframe are fused to form a host-side prior. This prior is attached to keyframe $T$ and subsequently used by the decoupled weighting scheme described in Sec.~\ref{sec:weight}.
 
Under ideal predictions of the proposed consistency network, short-term disturbances such as dynamic motion, occlusion, and abrupt illumination changes manifest as low quality in at least one of the adjacent frame pairs. Consequently, these structured sources of inconsistency are absorbed into the predicted quality, leaving only residual nominal variation. We model this remaining variation as an independent zero-mean Gaussian perturbation in the log-quality domain, analogous to the Gaussian treatment of inlier photometric residuals in RGBD-DSO~\cite{RGBD-DSO}.

Specifically, for a keyframe $T$, let $\mu^k(\mathbf{p})$ denote the latent host-side log-quality at pixel location $\mathbf{p}$. We model the two adjacent observations around $T$ as  measurements of this latent quantity:
\begin{equation}
\label{eq:log_quality_model}
\log Q^{k}_{T-1,T}(\mathbf{p}) = \mu^k(\mathbf{p}) + \eta_1,
\qquad
\log Q^{k}_{T,T+1}(\mathbf{p}) = \mu^k(\mathbf{p}) + \eta_2,
\end{equation}
where $\eta_1,\eta_2 \overset{\mathrm{i.i.d.}}{\sim} \mathcal{N}(0,\sigma^2)$ denote the nominal deviation of each adjacent observation from the host-side quantity.

Under this model, the maximum-likelihood estimate of $\mu^k(\mathbf{p})$ is the average of the two log-quality observations,
\begin{equation}
\label{eq:mle_logq}
\hat{\mu}^k(\mathbf{p})
=
\tfrac{1}{2}
\Bigl(
\log Q^{k}_{T-1,T}(\mathbf{p})
+
\log Q^{k}_{T,T+1}(\mathbf{p})
\Bigr),
\end{equation}
which yields the following absolute host-side quality prior in the original domain:
\begin{equation}
\label{eq:abs_prior}
Q^{k}_{\mathrm{abs}}(\mathbf{p};\,T)
=
\sqrt{
Q^{k}_{T-1,\,T}(\mathbf{p})
\;
Q^{k}_{T,\,T+1}(\mathbf{p})
}.
\end{equation}

Compared with using a single pairwise prediction, the bidirectional estimate reduces the variance from $\sigma^2$ to $\sigma^2/2$ and incorporates evidence from both temporal directions, so that disturbances confined to a single adjacent pair are reflected in the host-side prior rather than ignored. The effect of this design is evaluated against a variant that uses only the preceding pairwise quality (variant B) in Sec.~\ref{sec:ablation}.

\subsubsection{Decoupled Prior Weighting and Pose Estimation}
\label{sec:weight}

After pixels are selected for a host keyframe, the fused host-side priors are further used during direct pose estimation. In this stage, the prior no longer determines which pixels are selected but instead modulates how strongly each selected pixel contributes to the pose update. 

Given a host keyframe $T$ and an incoming frame $t$, the photometric residual at the $i$-th selected support pixel $\mathbf{p}_i$ is defined as
\begin{equation}
\label{eq:residual}
r_i(\boldsymbol{\xi})
=
I_T(\mathbf{p}_i)
-
I_t\!\left(
\pi\!\left(
\mathbf{T}(\boldsymbol{\xi})\mathbf{P}_i
\right)
\right),
\qquad
\mathbf{P}_i = D_T(\mathbf{p}_i)\,K^{-1}\bar{\mathbf{p}}_i,
\end{equation}
where $\mathbf{P}_i \in \mathbb{R}^3$ is the 3D point obtained by back-projecting $\mathbf{p}_i$ using the keyframe depth $D_T(\mathbf{p}_i)$, $\bar{\mathbf{p}}_i=[u_i,v_i,1]^\top$ is the homogeneous coordinate of $\mathbf{p}_i$, $\pi(\cdot)$ denotes perspective projection, $K$ is the intrinsic matrix, and $\mathbf{T}(\boldsymbol{\xi})=\exp(\boldsymbol{\xi})\in\mathrm{SE}(3)$ is the relative pose parameterized by $\boldsymbol{\xi}$. For clarity, we omit the affine brightness parameters $(a, b)$ used in 
RGBD-DSO~\cite{RGBD-DSO} to model exposure and gain variations across frames. They are jointly optimized with $\boldsymbol{\xi}$ in the same 8-DoF Gauss-Newton system but are not directly targeted by the proposed prior-weighting mechanism.

With the affine parameters omitted, we linearize Eq.~\ref{eq:residual} to first order with respect to the pose. Let the 6-DoF increment be partitioned as
\begin{equation}
\boldsymbol{\xi}
=
\begin{bmatrix}
\mathbf{t}^\top & \boldsymbol{\omega}^\top
\end{bmatrix}^\top,
\end{equation}
where $\mathbf{t}\in\mathbb{R}^3$ and $\boldsymbol{\omega}\in\mathbb{R}^3$ denote translation and rotation, respectively. For the warped 3D point $\mathbf{P}'_i=\mathbf{T}(\boldsymbol{\xi})\mathbf{P}_i$ and its image projection $\mathbf{p}'_i=\pi(\mathbf{P}'_i)$, let $\nabla I_t(\mathbf{p}'_i)=(d_x,d_y)^\top$ denote the target-image gradient at the projected pixel. The per-pixel Jacobian of the scalar residual with respect to $\boldsymbol{\xi}$ can then be partitioned as
\begin{equation}
\label{eq:jacobian_block}
\mathbf{J}_i
=
\begin{bmatrix}
\mathbf{J}_i^{\mathrm{tr}} &
\mathbf{J}_i^{\mathrm{rot}}
\end{bmatrix},
\end{equation}
where $\mathbf{J}_i^{\mathrm{tr}}$ and $\mathbf{J}_i^{\mathrm{rot}}$ denote the translational and rotational Jacobian blocks, respectively.

During coarse tracking, only the pose and affine parameters are optimized, and the keyframe depth is fixed in the linearization. Since perspective projection is invariant to a positive scaling of its argument, Eq.~\ref{eq:residual} can be rewritten as $\mathbf{p}'_i=\pi(\mathbf{R}K^{-1}\bar{\mathbf{p}}_i+\rho^{h}_i\mathbf{t})$ with $\rho^{h}_i=1/D_T(\mathbf{p}_i)$, where $\mathbf{R}$ and $\mathbf{t}$ are the rotation and translation of $\mathbf{T}(\boldsymbol{\xi})$. The host inverse depth thus appears only in product with $\mathbf{t}$. Consequently, differentiation with respect to $\mathbf{t}$ carries an inverse-depth factor $\rho_i$, whereas differentiation with respect to $\boldsymbol{\omega}$ does not. The Jacobian structure is therefore

\begin{equation}
\label{eq:jacobian_structure}
\mathbf{J}_i^{\mathrm{tr}}
=
\rho_i\,\mathbf{A}_i(d_x,d_y,u_i',v_i'),
\qquad
\mathbf{J}_i^{\mathrm{rot}}
=
\mathbf{B}_i(d_x,d_y,u_i',v_i'),
\end{equation}
where $(u_i',v_i')=\mathbf{p}'_i$, and $\mathbf{A}_i(\cdot)$ and $\mathbf{B}_i(\cdot)$ collect the standard
projection and motion terms of direct RGB-D tracking.

A relative depth error is equivalent to a proportional scaling of the translation vector $\mathbf{t}$ and can therefore be absorbed into the translation estimate while leaving the rotation estimate unaffected. In contrast, photometric error directly perturbs both the residuals and the image gradients, thereby affecting both the rotational and translational components of the optimization. 

Following the analysis above, we define the photometric and geometric weights from the absolute host-side quality priors as
\begin{equation}
\label{eq:weights}
w_p(\mathbf{p})
=
\sqrt{Q^{\mathrm{photo}}_{\mathrm{abs}}(\mathbf{p};\,T)+\epsilon},
\qquad
w_g(\mathbf{p})
=
\sqrt{Q^{\mathrm{geo}}_{\mathrm{abs}}(\mathbf{p};\,T)+\epsilon},
\end{equation}
where $\epsilon=10^{-4}$ ensures numerical stability. The square root is applied because the weights are squared in the information matrix; consequently, this choice makes the contribution of each pixel proportional to the quality prior $Q$ rather than to $Q^2$.

Each component of the Jacobian is scaled by the quality prior associated with the corresponding source of information:

\begin{equation}
\tilde{d}_x = w_p d_x,\qquad \tilde{d}_y = w_p d_y,\qquad \tilde{r}_i = w_p r_i,\qquad \tilde{\rho}_i = w_g \rho_i,
\end{equation}
yielding the weighted Jacobian

\begin{equation}
\label{eq:weighted_jacobian}
\tilde{\mathbf{J}}_i = \begin{bmatrix} w_p w_g\,\mathbf{J}_i^{\mathrm{tr}} & w_p\,\mathbf{J}_i^{\mathrm{rot}} \end{bmatrix}.
\end{equation}

The rotational block is scaled by $w_p$ alone, and the translational block by both. The weights are applied to $\mathbf{J}_i$ after it is formed, fixed for a keyframe, and reapplied to the recomputed $\mathbf{J}_i$ at each iteration. Since $w_g$ acts on part of the Jacobian only, $\tilde{\mathbf{J}}_i$ is not the Jacobian of any weighted residual. The resulting update therefore corresponds to a structured reweighting of the normal equations rather than to a conventional Gauss--Newton step derived from a single scalar objective. The weighted normal equations are given by

\begin{equation}
\label{eq:gauss_newton}
\mathbf{H} = \sum_i \tilde{\mathbf{J}}_i^\top \tilde{\mathbf{J}}_i,\qquad \mathbf{b} = \sum_i \tilde{\mathbf{J}}_i^\top \tilde{r}_i,\qquad \delta\boldsymbol{\xi} = -\mathbf{H}^{-1}\mathbf{b},
\end{equation}
followed by the pose update

\begin{equation}
\mathbf{T}\leftarrow \exp(\delta\boldsymbol{\xi})\,\mathbf{T}.
\end{equation}

Finally, the proposed decoupled weighting is applied only at the finest pyramid level ($l=0$), where the learned uncertainty maps are defined. Extending them to coarser pyramid levels would require cross-scale resampling, which may distort the predicted reliability structure through interpolation. We therefore retain the original RGBD-DSO weighting at coarser levels and apply the learned priors only in the final full-resolution alignment stage.

\subsection{Keyframe-based Sliding-Window Optimization}
\label{sec:optimization}

Following the RGBD-DSO, Con-DSO maintains a bounded window of active keyframes and jointly refines their poses via local optimization. Old keyframes are marginalized as they leave the window, preserving recent multi-view constraints at bounded cost. Since the proposed consistency priors do not modify this stage, they influence it only indirectly through the selected pixels and the prior-weighted tracking established in the preceding stages.

\section{EXPERIMENT AND RESULTS}
\label{sec:experiments}

In the experiments, we first inspect the learned uncertainty maps to verify whether the consistency network responds to typical challenging factors, and then evaluate Con-DSO on five public RGB-D benchmarks: ICL-NUIM~\cite{ICL}, RGB-D Scenes V2~\cite{V2}, TUM RGB-D~\cite{TUM}, BONN~\cite{BONN}, and OpenLORIS~\cite{Openloris}. These datasets cover controlled synthetic scenes and diverse real-world environments with dynamic objects, occlusions, illumination changes, and texture sparsity. The consistency network is trained only on TartanAir and is applied to all test datasets without fine-tuning. 

We compare Con-DSO against representative RGB-D VO/SLAM systems, including CVO, ACVO, and RGBD-DSO as direct odometry baselines, and ORB-SLAM3 and SR-SLAM as full-SLAM references with backend optimization and loop-closure capabilities. As Con-DSO is designed as an RGB-D direct odometry framework, CVO, ACVO, and RGBD-DSO serve as the primary baselines, whereas ORB-SLAM3 and SR-SLAM are included for reference and are not directly comparable in system functionality. We additionally compare against DROID-SLAM, a learning-based system with recurrent updates and global bundle adjustment over the entire sequence, as a reference from a different paradigm.

Performance is evaluated using Absolute Trajectory Error (ATE), Translational Relative Pose Error (T.RPE) and Rotational Relative Pose Error (R.RPE), all reported as RMSE using the evo toolkit. Due to the page limit, the main paper focuses on ATE as the primary global trajectory metric, while the comparison with DROID-SLAM and the complete T.RPE and R.RPE results are provided in the supplementary material. Experiments are conducted on Ubuntu 20.04 with an AMD Ryzen R7 CPU, an NVIDIA RTX 3080 GPU, and 32 GB RAM.

\subsection{Qualitative Analysis of Learned Uncertainty}
\label{sec:Learned Uncertainty}

\begin{figure*}[!t]
  \centering
  \includegraphics[width= 1.0\linewidth]{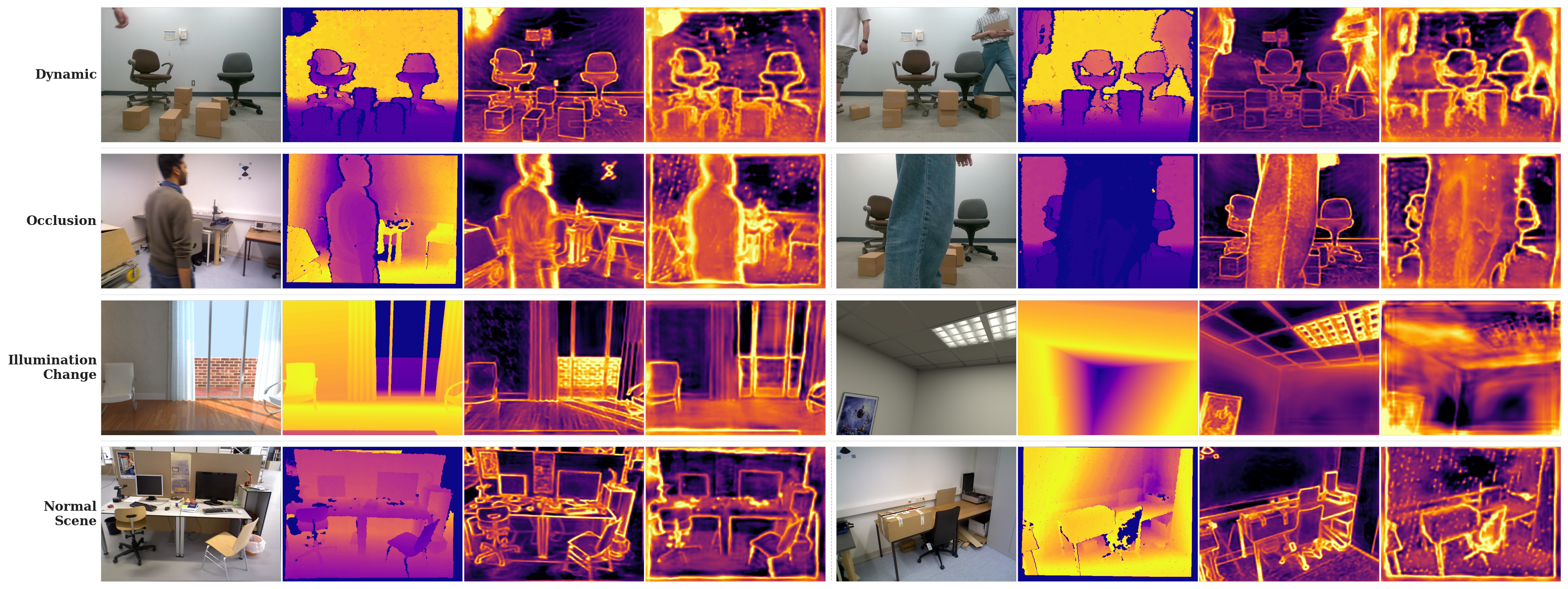}
  \caption{Qualitative visualization of learned photometric and geometric uncertainty under representative degradation factors. Brighter regions indicate higher predicted uncertainty.}
  \label{fig:demo}
\end{figure*}

Fig.~\ref{fig:demo} visualizes the predicted photometric and geometric uncertainty maps under representative conditions, where brighter regions indicate higher predicted uncertainty. In dynamic scenes, the photometric branch highlights moving regions, from a partially visible hand to full human bodies, while the geometric branch also responds to associated depth occlusion and discontinuities. Under occlusion, both dynamic and static occluders produce elevated geometric uncertainty along occlusion boundaries. For illumination changes, the photometric branch responds strongly to overexposed or reflective areas, such as sunlight entering through a balcony or bright ceiling lights. In normal static scenes, uncertainty is mainly concentrated around object boundaries and regions with unreliable depth. These results demonstrate that the learned consistency prior provides a unified response to diverse degradation factors without relying on semantic labels or designed rules.

We further evaluate the predicted uncertainty through sparsification analysis on all adjacent frame pairs from the five benchmark datasets. Following prior work, we use the Area Under the Sparsification Error (AUSE)~\cite{Ilg} and the Area Under the Random Gain (AURG)~\cite{Poggi}, which quantify, respectively, the gap between the predicted ranking and an oracle ranking derived from reference errors computed using ground-truth poses and sensor depth, and the improvement over random sample removal. Relative to the maximum achievable gain between random and oracle sparsification, the photometric and geometric branches attain average scores of 74.2\% and 75.4\%, respectively. Complete quantitative results and a detailed analysis are provided in the supplementary material.

\subsection{Main Quantitative Results}

\begin{table*}[!t]
\centering
\caption{ATE COMPARISON ON FIVE RGB-D BENCHMARKS}
\label{tab:ate_summary}
\small
\vspace{-2mm}
\setlength{\tabcolsep}{6.0pt}
\begin{tabular}{lcccccc|lcccccc}
\hline
\rule{0pt}{2.5ex}
\hspace{-1.2mm}\textbf{Seq.} 
& \textbf{CVO} & \textbf{ACVO} & \textbf{ORB3} & \textbf{SR} & \textbf{RGBD} & \textbf{Ours}
&
\hspace{0mm}\textbf{Seq.} 
& \textbf{CVO} & \textbf{ACVO} & \textbf{ORB3} & \textbf{SR} & \textbf{RGBD} & \textbf{Ours} \\
\hline
\rule{0pt}{2.5ex}

\hspace{-1.2mm}ICL\_01 
& 5.21 & 5.59 & 0.58 & 45.81 & \underline{0.27} & \textcolor{red}{\textbf{0.19}}
& w\_xyz 
& 1.644 & 1.666 & 0.513 & \textbf{0.015} & 0.119 & \textcolor{red}{\underline{0.039}} \\

ICL\_02 
& 1.43 & 0.88 & 8.42 & 9.68 & \underline{0.20} & \textcolor{red}{\textbf{0.19}}
& w\_rpy 
& 1.703 & 1.586 & 0.624 & \textbf{0.028} & 0.326 & \textcolor{red}{\underline{0.088}} \\

ICL\_03 
& 3.07 & 2.71 & 1.43 & 49.36 & \textbf{0.78} & \underline{0.79}
& w\_half 
& 1.218 & 1.134 & 0.270 & \textbf{0.022} & 0.239 & \textcolor{red}{\underline{0.139}} \\

ICL\_04 
& 21.62 & 22.52 & 1.60 & 77.97 & \underline{0.61} & \textcolor{red}{\textbf{0.26}}
& crowd 
& 2.610 & 2.568 & 1.192 & \textbf{0.147} & 1.332 & \textcolor{red}{\underline{0.558}} \\

ICL\_05 
& 6.94 & 6.08 & \underline{3.18} & \textbf{1.35} & 6.02 & \textcolor{red}{4.18}
& mov\_box 
& 0.614 & 0.591 & 0.241 & \underline{0.035} & 0.142 & \textcolor{red}{\textbf{0.027}} \\

ICL\_06 
& 4.88 & 4.18 & 73.06 & 91.33 & \textbf{0.92} & \underline{0.95}
& person
& 0.638 & 0.519 & 0.777 & \textbf{0.085} & 0.805 & \textcolor{red}{\underline{0.350}} \\

ICL\_07 
& 4.10 & 4.19 & 1.10 & 1.64 & \underline{0.83} & \textcolor{red}{\textbf{0.81}}
& sync
& 1.097 & 1.172 & 1.314 & 0.011 & \textbf{0.008} & \underline{0.009}  \\

ICL\_08 
& 10.30 & 9.56 & 1.51 & 1.52 & \textbf{0.87} & \underline{0.93}
& cafe 
& \underline{0.841} & \textbf{0.833} & 4.407 & 4.136 & 2.051 & 1.635 \\

V2\_01 
& 11.62 & 12.29 & \underline{1.28} & \textbf{1.22} & 9.94 & \textcolor{red}{1.52}
& office 
& 0.218 & 0.210 & \underline{0.066} & \textbf{0.065} & 0.266 & \textcolor{red}{0.072} \\

V2\_02 
& 10.67 & 12.54 & \underline{1.60} & \textbf{1.27} & 26.78 & \textcolor{red}{2.92}
& home 
& \underline{0.807} & \textbf{0.727} & 2.929 & 2.901 & x & 1.459 \\

V2\_03 
& 15.63 & 16.89 & \underline{1.40} & \textbf{1.27} & 22.33 & \textcolor{red}{5.70}
& market 
& 25.444 & 46.494 & 16.448 & 3.062 & \underline{2.939} & \textcolor{red}{\textbf{0.804}} \\

V2\_04 
& 13.69 & 14.22 & \underline{1.50} & \textbf{1.41} & 30.16 & \textcolor{red}{4.84}
& -- & -- & -- & -- & -- & -- & -- \\

\hline
\rule{0pt}{2.5ex}

\hspace{-1.2mm}\textbf{ICL Avg} 
& 7.19 & 6.96 & 11.36 & 34.83 & \underline{1.31} & \textcolor{red}{\textbf{1.04}}
& \textbf{TUM Avg}
& 1.521 & 1.462 & 0.469 & \textbf{0.022} & 0.228 & \textcolor{red}{\underline{0.089}} \\

\textbf{V2 Avg}
& 12.90 & 13.99 & \underline{1.45} & \textbf{1.29} & 22.30 & \textcolor{red}{3.75}
& \textbf{BONN Avg}
& 1.240 & 1.212 & 0.881 & \textbf{0.070} & 0.572 & \textcolor{red}{\underline{0.236}} \\

-- & -- & -- & -- & -- & -- & --
& \textbf{OL Avg}$^\dagger$
& 8.834 & 15.846 & 6.974 & 2.421 & \underline{1.752} & \textcolor{red}{\textbf{0.837}} \\

\hline
\end{tabular}
\vspace{0.2mm}
\begin{flushleft}
\hspace{0mm}\scriptsize \textbf{Note:} The best and second-best results in each row are highlighted in bold and underlined, respectively. \textcolor{red}{Values of Ours in red are the best among the direct methods (CVO, ACVO, RGBD-DSO, and Ours).} ORB3, SR, RGBD, and OL denote ORB-SLAM3, SR-SLAM, RGBD-DSO, and OpenLORIS, respectively. ICL-NUIM and RGB-D Scenes V2 are reported in cm, while TUM RGB-D, BONN, and OpenLORIS are reported in m. ``x'' denotes that RGBD-DSO tracks only 20\% of frames. $^\dagger$OpenLORIS average is computed over the three complete sequences used by all compared methods. Full T.RPE/R.RPE results are provided in the supplementary material.
\end{flushleft}
\end{table*}

Table~\ref{tab:ate_summary} reports the ATE on the five benchmarks, in centimeters for ICL-NUIM and RGB-D Scenes V2 and in meters otherwise. On the noise-free synthetic ICL-NUIM sequences (ICL\_01--ICL\_08), whose large weakly textured regions challenge feature tracking, the sparse direct methods remain the most accurate, and Con-DSO reduces the average ATE of RGBD-DSO from 1.31~cm to 1.04~cm. On RGB-D Scenes V2 (V2\_01--V2\_04), Kinect depth noise and invalid depth degrade RGBD-DSO to 22.30~cm, whereas Con-DSO reaches 3.75~cm, the same order of magnitude as SR-SLAM and ORB-SLAM3, which obtain the best results on these feature-rich tabletop scenes with loop closure and global bundle adjustment. On the dynamic TUM RGB-D (w\_xyz--w\_half) and BONN sequences (crowd--sync), SR-SLAM achieves the best averages through explicit dynamic detection, while Con-DSO clearly outperforms all other compared methods and obtains the lowest ATE on mov\_box, where the moving object is not a predefined semantic category. On OpenLORIS (cafe--market), the ranking depends on the dominant degradation: the point-cloud methods perform best in the small cafe and home scenes and the feature-based methods in the well-textured office, whereas Con-DSO achieves the best result on the large-scale market sequence (0.804~m) and thus the best average over the complete sequences.

Against DROID-SLAM, Con-DSO achieves lower ATE on 9 of the 23 evaluated sequences, including the large-scale market sequence, while DROID-SLAM, which performs recurrent correspondence refinement and confidence estimation together with sequence-level optimization, achieves lower ATE on the remaining 14 sequences. Complete per-sequence results are provided in the supplementary material.

\begin{figure*}[!t]
  \centering
  \includegraphics[width= 0.9\linewidth]{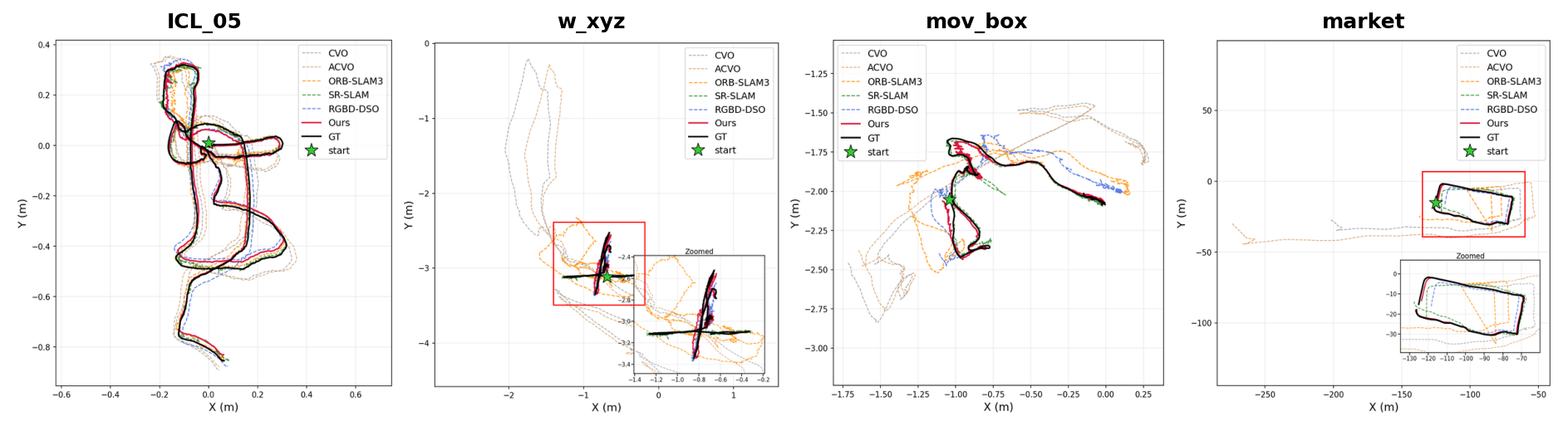}
  \caption{Representative trajectory comparisons on ICL\_05, TUM w\_xyz, BONN mov\_box, and OpenLORIS market. The green star denotes the start point. The black solid line denotes the ground truth, the red solid line denotes Con-DSO, and the dashed lines denote baseline methods: gray for CVO, brown for ACVO, yellow for ORB-SLAM3, green for SR-SLAM, and blue for RGBD-DSO.}
  \label{fig:traj_representative}
\end{figure*}

Fig.~\ref{fig:traj_representative} provides representative trajectory comparisons. Con-DSO remains close to the ground truth across weak-texture, dynamic, manipulated-object, and large-scale scenes, whereas direct baselines often show accumulated drift and full-SLAM references may still degrade. Complete trajectory visualizations are included in the supplementary material.

In summary, Con-DSO attains the lowest ATE among the direct methods on all five benchmarks. Its improvement over the direct baseline is smallest on the noise-free synthetic ICL-NUIM ($1.26\times$) and largest on real data, up to $5.95\times$ on RGB-D Scenes V2, where the sensor depth is noisy. It also attains the lowest ATE overall on the OpenLORIS average and on market, the largest real-world sequence, with $0.804$~m against $3.062$~m for the challenge-aware full-SLAM system SR-SLAM.

\subsection{Ablation Study}
\label{sec:ablation}

To analyze the contribution of each component, we conduct ablation studies on all datasets (Table~\ref{tab:ablation_ate}). Variant A additionally applies the geometric weight to the rotational block. Variant B omits the bidirectional fusion and uses the pairwise quality of the preceding pair alone. In variants C1 and C2, only the photometric or only the geometric branch takes effect. Since the photometric branch also drives pixel selection, C2 reverts to gradient-based selection. Variants P and W apply the prior to pixel selection only and to weighting only, respectively.

\begin{table}[!htbp]
\centering
\caption{ABLATION STUDY OF ATE RMSE}
\label{tab:ablation_ate}
\small
\vspace{-2mm}
\setlength{\tabcolsep}{2.5pt}
\begin{tabular}{llccccc}
\toprule
\rule{0pt}{2.5ex}
 & \textbf{Variant} & \textbf{ICL} & \textbf{V2} & \textbf{TUM} & \textbf{BONN} & \textbf{LORIS}$^\dagger$ \\
 & & (cm) & (cm) & (m) & (m) & (m) \\
\midrule
\rule{0pt}{2.5ex}
-- & RGBD-DSO (no prior)       & 1.31 & 22.30 & 0.228 & 0.572 & 1.752 \\
\midrule
A  & Shared weighting          & 1.64 & 15.15 & 0.152 & 0.408 & 1.229 \\
B  & Preceding-pair prior        & 1.53 & 20.13 & 0.134 & 0.558 & 0.996 \\
C1 & Photometric branch only   & 1.23 & 15.50 & 0.162 & 0.424 & 0.988 \\
C2 & Geometric branch only     & 1.34 & 16.83 & 0.117 & 0.630 & 1.144 \\
P & Selection only           & 1.59 & 18.88 & 0.129 & 0.401 & 1.379 \\
W & Weighting only           & 1.15 & 13.60 & 0.111 & 0.353 & 1.302 \\
\midrule
-- & Full (Con-DSO)            & \textbf{1.04} & \textbf{3.75} & \textbf{0.089} & \textbf{0.236} & \textbf{0.837} \\
\bottomrule
\end{tabular}
\vspace{0.2mm}
\begin{flushleft}
\scriptsize \textbf{Note:} $^\dagger$OpenLORIS average is computed over the three complete sequences.
\end{flushleft}
\end{table}

Applying the geometric weight to the rotational block (A) degrades all five datasets, most severely on RGB-D Scenes V2, where depth errors cause the geometric weight to suppress valid rotational constraints. Using only the preceding pairwise quality (B) loses most of the improvement on RGB-D Scenes V2 and BONN. Retaining only the photometric branch (C1) yields improvements over RGBD-DSO across all five benchmarks. Using only the geometric branch (C2) results in lower accuracy than RGBD-DSO on ICL-NUIM and BONN. Applying the prior to pixel selection only (P) degrades ICL-NUIM (1.31 to 1.59\,cm), while applying it to weighting only (W) improves on RGBD-DSO on all five benchmarks. Overall, the full Con-DSO formulation achieves the lowest ATE on every benchmark, indicating that each component contributes and that the components are complementary when combined.

\subsection{Runtime Analysis}

Con-DSO runs at 54.69--101.04~ms/frame across the evaluated datasets. Compared with RGBD-DSO, the additional overhead mainly comes from network inference and host-side prior construction, while pixel ranking and decoupled weighting introduce only minor cost. Although slower than ORB-SLAM3 and RGBD-DSO, Con-DSO remains faster than dense point-cloud registration methods such as CVO and ACVO, and achieves runtime comparable to SR-SLAM on dynamic benchmarks. Detailed runtime breakdowns are provided in the supplementary material. 

\subsection{Threshold Analysis}
\label{sec:threshold analysis}

To assess the robustness of fixed parameter settings across datasets, we further examine whether fixed settings transfer across benchmarks by normalizing the average ATE of the four direct methods, each run under a single setting on all five benchmarks, by their mean on each benchmark. The relative standing of CVO, ACVO, and RGBD-DSO varies considerably across benchmarks, whereas Con-DSO, which introduces no additional thresholds beyond those of RGBD-DSO, remains the most accurate on all five. The normalized results and a detailed analysis are given in the supplementary material.

\section{Conclusion}

In this paper, we proposed Con-DSO, a consistency-aware RGB-D direct sparse odometry framework for challenging environments. Con-DSO represents consistency violations from diverse sources in a unified manner as photometric and depth-geometric uncertainty learned from temporally adjacent RGB-D frame pairs. The resulting pairwise quality guides uncertainty-aware support-pixel selection, and its bidirectional fusion into a host-side quality prior drives decoupled photometric-geometric weighting during pose estimation. Experiments on five public RGB-D datasets show that this learned prior improves the accuracy and robustness of direct RGB-D odometry across synthetic and real-world scenes at a practical runtime.

Despite these advantages, Con-DSO has several limitations. First, as a unified prior not designed for any particular challenge, it may fall short of methods specialized for specific conditions. Second, the prior is built from short-horizon adjacent pairs and used only for local keyframe-based tracking. Extending it to a temporally persistent~\cite{Khronos} quality model for global optimization and map maintenance is a promising direction. Finally, the prior is predicted once and held fixed during optimization. A natural extension would be to update the prior recurrently during optimization, analogous to the iterative confidence refinement employed by DROID-SLAM~\cite{DROID}, which we leave for future work.


 





\end{document}